\documentclass[runningheads]{llncs}
\usepackage{graphicx}
\usepackage{tikz}
\usepackage{comment}
\usepackage{amsmath,amssymb} % define this before the line numbering.
\usepackage{color}

\usepackage[accsupp]{axessibility}  % Improves PDF readability for those with disabilities.

\usepackage[width=122mm,left=12mm,paperwidth=146mm,height=193mm,top=12mm,paperheight=217mm]{geometry}

\usepackage{booktabs}
\usepackage{bm}
\usepackage{mathtools}
\usepackage{algpseudocode}
\usepackage{algorithm}
\usepackage{siunitx}
\usepackage{multirow}
\usepackage{enumitem}
\usepackage{capt-of}
\usepackage{xcolor}
\usepackage{diagbox}
\usepackage{subfigure}

\usepackage[pagebackref,breaklinks,colorlinks]{hyperref}

\makeatletter

\usepackage[capitalize]{cleveref}
\crefname{section}{Sec.}{Secs.}
\Crefname{section}{Section}{Sections}
\Crefname{table}{Table}{Tables}
\crefname{table}{Tab.}{Tabs.}

\begin{document}
% \linenumbers
\pagestyle{headings}
\mainmatter
\def\ECCVSubNumber{6210}  % Insert your submission number here

\title{Lipschitz Continuity Retained Binary Neural Network} % Replace with your title

\begin{comment}
\titlerunning{ECCV-22 submission ID \ECCVSubNumber} 
\authorrunning{ECCV-22 submission ID \ECCVSubNumber} 
\author{Yuzhang Shang}
\institute{Paper ID \ECCVSubNumber}
\end{comment}
%******************

% USER DEFINED
\graphicspath{ {images/} }

\algnewcommand\algorithmiclocalize{\textbf{Localize neuron and update model:}}
\algnewcommand\Localize{\item[\algorithmiclocalize]}
\algnewcommand\algorithmiccross{\textbf{Cross-section plane determination:}}
\algnewcommand\Cross{\item[\algorithmiccross]}
\algnewcommand\algorithmicbranchc{\textbf{Bifurcation candidates detection:}}
\algnewcommand\Branchc{\item[\algorithmicbranchc]}

\newcommand{\fft}[1]{\bm{\mathbf{\hat{#1}}}}
\newcommand{\norm}[1]{\vert \vert{#1}\vert \vert^2}
\newcommand{\normd}[1]{\vert \vert{#1}\vert \vert^2_2}
\newcommand{\diag}[1]{\text{diag}}
\newcommand{\conj}[1]{\text{conj}}

\newcommand{\syz}[1]{\textcolor{blue}{#1}}

\renewcommand{\vec}[1]{\boldsymbol{#1}}

% \begin{comment}
\titlerunning{}
\author{Yuzhang Shang\inst{1}\and
Dan Xu\inst{2} \and Bin Duan\inst{1}\and \\ Ziliang Zong\inst{3}\and Liqiang Nie\inst{4}\and
Yan Yan\inst{1}\thanks{Corresponding author.}}
\authorrunning{Y. Shang et al.}
\institute{Illinois Institute of Technology, USA \and
Hong Kong University of Science and Technology, Hong Kong \and
Texas State University, USA \and Harbin Institute of Technology, Shenzhen, China\\
 \email{\{yshang4, bduan2\}@hawk.iit.edu, danxu@cse.ust.hk, ziliang@txstate.edu, nieliqiang@gmail.com, and yyan34@iit.edu}
}
% \institute{Department of Computer Science, Illinois Institute of Technology, USA \and
% Department of Computer Science and Engineering, HKUST, Hong Kong \and
% Department of Computer Science, Texas State University, USA \and School of Computer Science and Technology, Harbin Institute of Technology, Shenzhen, China\\
%  \email{yshang4@hawk.iit.edu, danxu@cse.ust.hk, bduan2@hawk.iit.edu,  ziliang@txstate.edu, nieliqiang@gmail.com, and yyan34@iit.edu}
% }

%******************
\maketitle

\begin{abstract}
\label{sec:abstract}
Relying on the premise that the performance of a binary neural network can be largely restored with eliminated quantization error between full-precision weight vectors and their corresponding binary vectors, existing works of network binarization frequently adopt the idea of model robustness to reach the aforementioned objective. However, robustness remains to be an ill-defined concept without solid theoretical support. In this work, we introduce the Lipschitz continuity, a well-defined functional property, as the rigorous criteria to define the model robustness for BNN. We then propose to retain the Lipschitz continuity as a regularization term to improve the model robustness. Particularly, while the popular Lipschitz-involved regularization methods often collapse in BNN due to its extreme sparsity, we design the Retention Matrices to approximate spectral norms of the targeted weight matrices, which can be deployed as the approximation for the Lipschitz constant of BNNs without the exact Lipschitz constant computation (NP-hard). Our experiments prove that our BNN-specific regularization method can effectively enhance the robustness of BNN (testified on ImageNet-C), achieving SoTA on CIFAR10 and ImageNet. Our code is available at \url{https://github.com/42Shawn/LCR_BNN}.


\end{abstract}
\section{Introduction}
\label{sec:intro}
\sloppy
Recently, Deep Neural Networks achieve significant accomplishment in computer vision tasks such as image classification~\cite{krizhevsky2012imagenet} and object detection~\cite{ren2015faster,lecun2015deep}. However, their inference-cumbersome problem hinders their broader implementations. To develop deep models in resource-constrained edge devices, researchers propose several neural network compression paradigms, \textit{e.g.,} knowledge distillation~\cite{hinton2015distilling,heo2019comprehensive}, network pruning~\cite{lecun1989optimal,han2015deep} and network quantization~\cite{hubara2016binarized,qin2020forward}. Among the network quantization methods, the network binarization~\cite{hubara2016binarized} stands out, as it extremely quantizes weights and activations (\textit{i.e.}~intermediate feature maps) to $\pm 1$. Under this framework, the full-precision (FP) network is compressed 32$\times$ more, and the time-consuming inner-product operations are replaced with the efficient Xnor-bitcount operations.

However, BNNs can hardly achieve comparable performance to the original models due to the loss of FP weights and activations. A major reason for the performance drop is that the inferior robustness comes from the error amplification effect, where the binarization operation degrades the distance induced by amplified noise~\cite{lin2019defensive}. The destructive manner of $\textit{sgn}(\cdot)$ severely corrupts the robustness of the BNN, and thus undermines their representation capacity ~\cite{bulat2019xnor,he2020proxybnn,liu2020reactnet}.

As some theoretical works validated, robustness is a significant property for functions (neural networks in our context), which further influences their generalization ability~\cite{luxburg2004distance,bartlett2017spectrally}. In the above-mentioned binarization works, researchers investigate the effectiveness of their methods via the ill-defined concepts of function robustness without solid theoretical support, such as observing the visualized distributions of weights and activations~\cite{he2020proxybnn,lin2020rotated,liu2020reactnet,lin2019defensive}. However, they rarely introduced the well-defined mathematical property, Lipschitz continuity, for measuring the robustness of functions into BNN. Lipschitz continuity has been proven to be a powerful and strict tool for systematically analyzing deep learning models. For instance, Miyato~\textit{et. al.} propose the well-known Spectral Normalization~\cite{yoshida2017spectral,miyato2018spectral} utilizing the Lipschitz constant to regularize network training, which is initially designed for GAN and then extended to other network architectures, achieving great success~\cite{neyshabur2017exploring}; Lin~\textit{et. al.}~\cite{lin2019defensive} design a Lipschitz-based regularization method for network (low-bit) quantization, and testify that Lipschitz continuity is significantly related to the robustness of the low-bit network. But simply bridging those existing Lipschitz-based regularization methods with the binary neural networks (1-bit) is sub-optimal, as the exclusive property of BNN,~\textit{e.g.}, the extreme sparsity of binary weight matrix ~\cite{hubara2016binarized} impedes calculating the singular values, which is the core module in those Lipschitz-involved methods.

To tackle this problem, we analyze the association between the structures and the Lipschitz constant of BNN. Motivated by this analysis, we design a new approach to effectively retain the Lipschitz constant of BNNs and make it close to the Lipschitz constant of its latent FP counterpart. Particularly, we develop a Lipschitz Continuity Retention Matrix ($\mathbf{RM}$) for each block and calculate the spectral norm of $\mathbf{RM}$ via the iterative power method to avoid the high complexity of calculating exact Lipschitz constants. It is worth to note that the designed loss function for retaining the Lipschitz continuity of BNNs is differentiable \emph{w.r.t.} the binary weights.
% Note that our BNN-specific Lipschitz regularization term takes full advantage of the distinct structure of BNN where binary and FP activations exist in the same forward pass and thus avoid the collapse of previous Lipschitz-involved algorithms in processing sparse 1-bit weight matrices.

\noindent Overall, the contributions of this paper are three-fold:
\begin{itemize}[leftmargin=*, topsep=0pt, partopsep=0pt, itemsep=1pt]
    \item We propose a novel network binarization framework, named as ~\textbf{L}ipschitz \textbf{C}ontinuity \textbf{R}atined Binary Neural Network  (\textbf{\emph{LCR}}-BNN), to enhance the robustness of binary network optimization process. To the best of our knowledge, we are the first on exploring the Lipschitz continuity to enhance the representation capacity of BNNs;
    \item We devise a Lipschitz Continuity Retention Matrix to approximate the Lipschitz constant with activations (instead of directly using weights as SN~\cite{miyato2018spectral} and DQ~\cite{lin2019defensive} devised) of networks in the BNN forward pass; 
    \item By adding our designed regularization term on the existing state-of-the-art methods, we observe the enhanced robustness are validated on ImageNet-C and promising accuracy improvement on CIAFR and ImageNet datasets. 
\end{itemize}
\section{Related Work}
\label{related}
\sloppy
\subsection{Network Binarization} In the pioneer art of BNNs, Hubara \textit{et. al.}~\cite{hubara2016binarized} quantize weights and activations to $\pm 1$ via sign function. Due to the non-differentiability of the sign function, the straight-through estimator (STE)~\cite{bengio2013estimating} is introduced for approximating the derivative of the sign function. Inspired by this archetype, numerous researchers dig into the field of BNNs and propose their modules to improve the performance of BNNs. For instance, Rastegari \textit{et. al.}~\cite{rastegari2016xnor} reveal that the quantization error between the FP weights and corresponding binarized weights is one of the obstacles degrading the representation capabilities of BNNs. Then they propose to introduce a scaling factor calculated by the L1-norm for both weights and activation functions to minimize the quantization error. XNOR++~\cite{bulat2019xnor} absorbs the idea of scaling factor and proposes learning both spatial and channel-wise scaling factors to improve performance. Furthermore, Bi-Real~\cite{liu2020bi} proposes double residual connections with full-precision downsampling layers to lessen the information loss. ProxyBNN~\cite{he2020proxybnn} designs a proxy matrix as a basis of the latent parameter space to guide the alignment of the weights with different bits by recovering the smoothness of BNNs.
Those methods try to lessen the quantization error and investigate the effectiveness from the perspective of model smoothness (normally via visualizing the distribution of weights). A more detailed presentation and history of BNNs can be found in the Survey~\cite{qin2020binary}.

However, none of them take the functional property, Lipschitz continuity, into consideration, which is a well-developed mathematical tool to study the robustness of functions. Bridging Lipschitz continuity with BNNs, we propose to retain the Lipschitz continuity of BNNs, which can serve as a regularization term and further improve the performance of BNNs by strengthening their robustness.

\subsection{Lipschitz Continuity in Neural Networks}
The Lipschitz constant is an upper bound of the ratio between input perturbation and output variation within a given distance. It is a well-defined metric to quantify the robustness of neural networks to small perturbations~\cite{scaman2018lipschitz}. Also, the Lipschitz constant $\Vert f \Vert_{Lip}$ can be regarded as a functional norm to measure the Lipschitz continuity of given functions. Due to its property, the Lipschitz constant is the primary concept to measure the robustness of functions~\cite{bartlett2017spectrally,luxburg2004distance,neyshabur2017exploring}. 
In the deep learning era, previous theoretical arts~\cite{virmaux2018lipschitz,neyshabur2017exploring} disclose the regularity of deep networks via Lipschitz continuity. Lipschitz continuity is widely introduced into many deep learning topics for achieving the SoTA performance~\cite{miyato2018spectral,yoshida2017spectral,shang2021lipschitz,zhang2021towards}. For example, in image synthesis, Miyato~\textit{et. al.}~\cite{miyato2018spectral,yoshida2017spectral} devise spectral normalization to constrain the Lipschitz constant of the discriminator for optimizing a generative adversarial network, acting as a regularization term to smooth the discriminator function; in knowledge distillation, Shang \textit{et. al.}~\cite{shang2021lipschitz} propose to utilize the Lipschitz constant as a form of knowledge to supervise the training process of student network; in neural network architecture design, Zhang \textit{et. al.}~\cite{zhang2021towards} propose a novel $L_{\infty}$-dist network using naturally 1-Lipschitz functions as neurons.

The works above highlight the significance of Lipschitz constant in expressiveness and robustness of deep models. Particularly, retaining Lipschitz continuity at an appropriate level is proven to be an effective technique for enhancing the model robustness. Therefore, the functional information of neural networks, Lipschitz constant, should be introduced into network binarization to fill the robustness gap between BNN and its real-valued counterpart. 

\noindent\textbf{Relation to Spectral Normalization (SN)~\cite{miyato2018spectral}.} We empirically implement the SN in BNN but fail. By analyzing the failure of the implementation, we conclude that the SN is not suitable for BNNs. The reasons are: 
(i) One of the key modules in SN is spectral norm computation based on singular value calculatiuon, which is directly implemented on the weight matrix (\textit{e.g.}, the matrices of convolutional and linear layers). But the binarization enforcing the FP weight into 1 or -1 makes the weight matrix extremely sparse. Thus, applying the existing algorithm to binary matrices collapses. (ii) In contrast to normal networks, the forward and backward passes of BNN are more complex, \textit{e.g.}, FP weights (after backpropagation) and binary weights (after binarization) exist in the same training iteration. This complexity problem impedes broader implementations of SN on BNNs as the number of structures in a BNN exceeds the number in a normal network. To tackle those problems, we propose a novel Lipschitz regularization technique targeted to train BNNs. We elaborate more technical comparisons between our method and SN in the following Section~\ref{sec:difference}. 

\section{Lipschitz Continuity Retention for BNNs}
\label{sec:method}
\subsection{Preliminaries}
\label{sec:pre}
We first define a general neural network with $L$ fully-connected layers (without bias term for simplification). This network $f(\mathbf{x})$ can be denoted as:
\begin{equation}
    f(\mathbf{W}^1,\cdots,\mathbf{W}^L;\mathbf{x}) = (\mathbf{W}^{L}\cdot\sigma\cdot \mathbf{W}^{L-1}\cdot \cdots \cdot\sigma\cdot \mathbf{W}^{1})(\mathbf{x}),
    \label{eq:1}
\end{equation}
where $\mathbf{x}$ is the input sample and $\mathbf{W}^{k}\in \mathbb{R}^{d_{k-1}\times d_{k}} (k=1,...,L-1)$ stands for the weight matrix connecting the $(k-1)$-th and the $k$-th layer, with $d_{k-1}$ and $d_{k}$ representing the sizes of the input and output of the $k$-th network layer, respectively. The $\sigma(\cdot)$ function performs element-wise activation for the activations. 
% Notably, for a convolution layer with the input map of $m$ channels and the output map of $n$ channels, and the size of the kernel $w\times h$, it results in $m\times n\times w\times h$ parameters. We can re-arrange the parameters to a weight matrix of size $n\times (m\times h\times w)$, such that this convolution layer can also operate in the same way as the other fully-connected layers do. Hence, it is sufficient to consider networks with the fully-connected layers.

\noindent \textbf{Binary Neural Networks.} Here, we revisit the general gradient-based method in~\cite{courbariaux2015binaryconnect}, which maintains full-precision latent variables $\mathbf{W}_F$ for gradient updates, and the $k$-th weight matrix $\mathbf{W}_F^k$ is binarized into $\pm 1$ binary weight matrix $\mathbf{W}_B^k$ by a binarize function (normally $\textit{sgn}(\cdot)$) as $\mathbf{W}_B^k = \textit{sgn}(\mathbf{W}_F^k)$. Then the activation map of the $k$-th layer is produced by $\mathbf{A}^{k} = \mathbf{W}_B^k \mathbf{A}^{k-1}$, and a whole forward pass of binarization is performed by iterating this process for $L$ times.

\noindent \textbf{Lipschitz Constant (Definition 1).} A function $g : \mathbb{R}^{n} \longmapsto \mathbb{R}^{m}$ is called Lipschitz continuous if there exists a constant $L$ such that:
\begin{equation}
    \forall \mathbf{x,y}  \in  \mathbb{R}^{n}, \Vert g(\mathbf{x}) - g(\mathbf{y})\Vert_2 \leq L\Vert \mathbf{x} - \mathbf{y}\Vert_2,
    \label{definition:lip}
\end{equation}
where $\mathbf{x,y}$ represent two random inputs of the function $g$. The smallest $L$ holding the inequality is the Lipschitz constant of function $g$, denoted as $\Vert g \Vert_{Lip}$. By Definition 1, $\Vert \cdot \Vert_{Lip}$ can upper bound of the ratio between input perturbation and output variation within a given distance (generally L2 norm), and thus it is naturally considered as a metric to evaluate the robustness of neural networks~\cite{scaman2018lipschitz,rosca2020case,shang2021lipschitz}.

In the following section, we propose our Lipschitz Continuity Retention Procedure (Sec.~\ref{sec:3.2}), where the a BNN is enforced to close to its FP counterpart in term of Lipschitz constant. In addition, we introduce the proposed loss function and gradient approximation for optimizing the binary network (Sec.~\ref{sec:3.3}). Finally, we discuss the relation between \emph{LCR} and Lipschitz continuity, and compare our method to the well-known Spectral Normalization~\cite{miyato2018spectral} (Sec.~\ref{sec:difference}).

\subsection{Lipschitz Continuity Retention Procedure}
\label{sec:3.2}
\begin{figure}[!t]
  \begin{center}
    \includegraphics[width=0.97\textwidth]{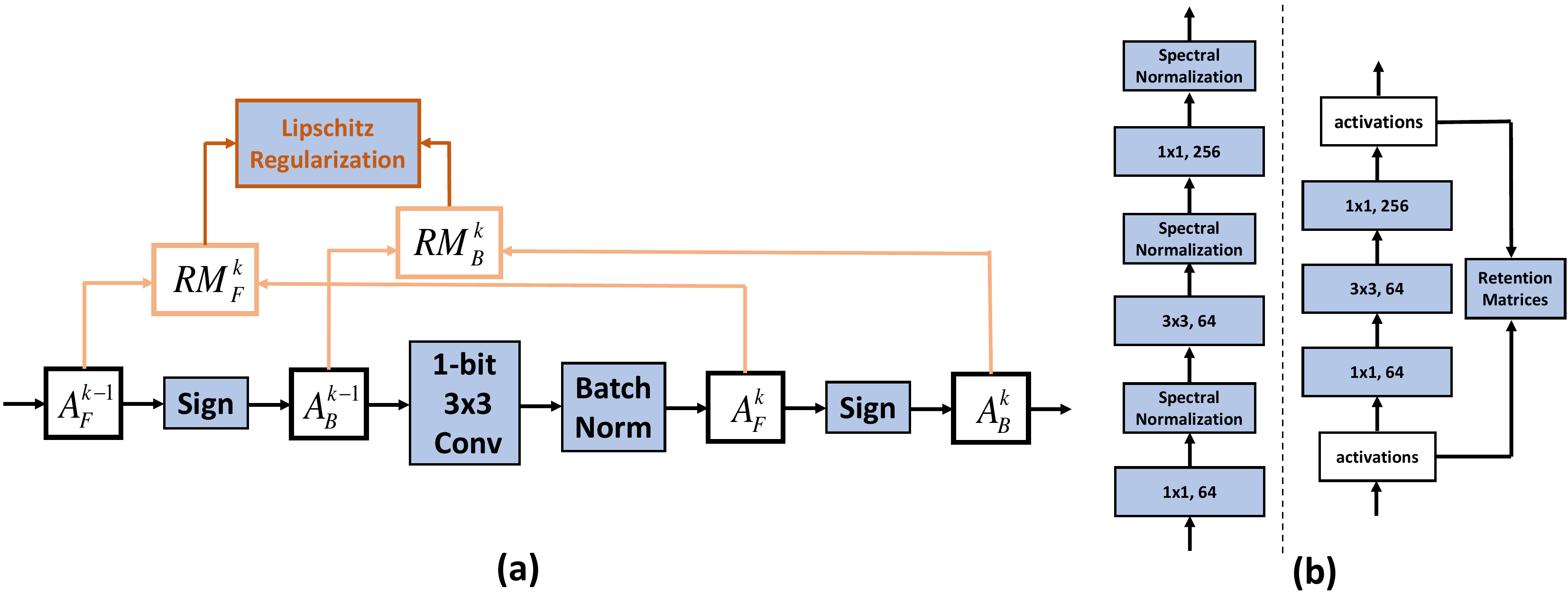}
  \end{center}
  \vspace{-0.3in}
  \caption{(\textbf{a}) An overview of our Lipschitz regularization for a binary convolutional layer: regularizing the BNN via aligning the Lipschitz constants of binary network and its latent full-precision counterpart is the goal of our work. To reach this goal, the input and output activations of the $k$-th layer compose the Retention Matrix ($\mathbf{RM}^k$) for approximating the Lipschitz constant of this layer. $\mathbf{RM}^k_F$ and $\mathbf{RM}^k_B$ are then used to calculate the Lipschitz constant of this layer (the validation of this approximation is elaborated in~\ref{sec:3.2}). Finally, the Lipschitz continuity of the BNN is retained under a regularization module. (\textbf{b}) Difference between Spectral Normalization (Left) and \emph{LCR} (Right). More details are discussed in~\ref{sec:difference}.}
\label{fig:pipeline}
\vspace{-0.15in}
\end{figure}

% Due to the existance of the Lipschitz continuity noise, we propose to minimize the distance between the Lipschitz constants of binary and full-precision networks.
We aim to retain the Lipschitz constants in an appropriate level. In practice, we need to pull $\Vert f_B\Vert_{Lip}$ and $\Vert f_F\Vert_{Lip}$ closely to stabilize the Lipschitz constant of the BNNs.
However, it is NP-hard to compute the exact Lipschitz constant of neural networks~\cite{virmaux2018lipschitz}, especially involving the binarization process. 
To solve this problem, we propose to bypass the exact Lipschitz constant computation by introducing a sequence of Retention Matrices produced by the adjacent activations, and then compute their spectral norms via power iteration method to form a LCR loss for retaining the Lipschitz continuity of the BNN as demonstrated in Figure~\ref{fig:pipeline}.

\noindent \textbf{Lipschitz constant of neural networks.} We fragment an affine function for the $k$-th layer with weight matrix $\mathbf{W}^k$, $f^k(\cdot)$ mapping $\mathbf{a}^{k-1} \longmapsto \mathbf{a}^{k}$, in which $\mathbf{a}^{k-1} \in \mathbb{R}^{d_{k-1}}$ and $\mathbf{a}^{k} \in \mathbb{R}^{d_{k}}$ are the activations produced from the $(k-1)$-th and the $k$-th layer, respectively.
Based on Lemma 1 in the Supplemental Materials, $\Vert f^k\Vert_{Lip}= {\sup}_{\mathbf{a}} \Vert \nabla \mathbf{W}^k(\mathbf{a}) \Vert_{SN}$, where $ \Vert \cdot \Vert_{SN}$ is the matrix spectral norm formally defined as:
\begin{equation}
\label{eq:3}
    \Vert \mathbf{W}^k \Vert_{SN} \triangleq \max \limits_{\mathbf{x}:\mathbf{x}\neq \mathbf{0}} \frac{\Vert \mathbf{W}^k \mathbf{x} \Vert_2}{\Vert \mathbf{x} \Vert_2} = \max \limits_{\Vert\mathbf{x}\Vert_2 \leq  1}{\Vert \mathbf{W}^k \mathbf{x} \Vert_2},
\end{equation}
where the spectral norm of the matrix $\mathbf{W}$ is equivalent to its largest singular value. Thus, for the $f^k$, based on Lemma 2 in the Supplemental Materials, its Lipschitz constant can be derived as:
\begin{equation}
    \Vert \mathbf{W}^k\Vert_{Lip} = {\sup}_{\mathbf{a}} \Vert \nabla \mathbf{W}^k(\mathbf{a}) \Vert_{SN} = \Vert \mathbf{W}^k \Vert_{SN}.
\end{equation}

Moreover, as for the most functional structures in neural network such as ReLU, Tanh, Sigmoid, Sign, batch normalization and other pooling layers, they all have simple and explicit Lipschitz constants~\cite{goodfellow2016deep,miyato2018spectral,shang2021lipschitz}. Note that for the sign function in BNN, though it is not theoretically differentiable, it still has an explicit Lipschitz constant as its derivative is numerically approximated by HardTanh function~\cite{bengio2013estimating}. This fixed Lipschitz constant property renders our derivation to be applicable to most network architectures, such as binary ResNet~\cite{he2016deep,hubara2016binarized} and variant binary ResNet~\cite{liu2020reactnet,bulat2020bats}.

By the inequality of norm, \textit{i.e.} $\Vert \mathbf{W}^k \cdot \mathbf{W}^{k+1} \Vert_{Lip} \leq \Vert \mathbf{W}^k \Vert_{Lip}\cdot \Vert \mathbf{W}^{k+1} \Vert_{Lip}$, we obtain the following upper bound of the Lipschitz constant of network $f$, \textit{i.e.,}
\begin{equation}
\begin{split}
    \Vert f \Vert_{Lip} \leq \Vert \mathbf{W}^L \Vert_{Lip} \cdot \Vert \sigma \Vert_{Lip} \cdots \cdot \Vert \mathbf{W}^1 \Vert_{Lip} 
    = \prod_{k=1}^{L}\Vert \mathbf{W}^{k}\Vert_{SN}.
\end{split}
\label{eq:7}
\end{equation}
In this way, we can retain the Lipschitz constant through maintaining a sequence of spectral norms of intermediate layers in the network.

\noindent \textbf{Construction of Lipschitz Continuity Retention Matrix.} We now aim to design a novel optimization loss to retain Lipschitz continuity by narrowing the distance between the spectral norms of corresponding weights of full-precision and binary networks. Moreover, we need to compute the spectral norm of binarized weight matrices. Nevertheless, it is inaccessible to calculate the spectral norm of the binary weight matrix $\mathbf{W}^k_B$ in BNNs by popular SVD-based methods~\cite{aharon2006k}. 
Therefore, we design the Lipschitz Continuity Retention Matrix ($\mathbf{RM}$) to bypass the complex calculation of the spectral norm of $\mathbf{W}_B^k$. Approaching the final goal through the bridge of the Retention Matrix allows feasible computation to retain the Lipschitz constant and facilitates its further use as a loss function.

For training data with a batch size of $N$, we have a batch of corresponding activations after a forward process for the ($k$-1)-th layer as
\begin{equation}
    \mathbf{A}^{k-1} = (\mathbf{a}^{k-1}_1,\cdots,\mathbf{a}^{k-1}_n) \in \mathbb{R}^{{d_{k-1}}\times N},
\end{equation}
where $\mathbf{W}^k \mathbf{A}^{k-1}= \mathbf{A}^{k}$ for each $k \in \{1,\dots,L-1\}$.

Studies about similarity of activations illustrate that for well-trained networks, their batch of activations in the same layer (\textit{i.e.}~$\{\mathbf{a}^{k-1}_i\}, i \in \{1,\dots,n\}$) have strong mutual linear independence. We formalize the independence of the activations as follows:
\begin{equation}
\begin{split}
    (\mathbf{a}^{k-1}_i)^{\mathsf{T}} \mathbf{a}^{k-1}_j \approx 0&,~~~~\forall i\neq j\in\{1,\cdots,N\},\\
    (\mathbf{a}^{k-1}_i)^{\mathsf{T}} \mathbf{a}^{k-1}_i \neq 0&,~~~~\forall i \in\{1,\cdots,N\}.
\end{split}
\label{eq:9}
\end{equation}
We also empirically and theoretically discuss the validation of this assumption in the Sec.~\ref{sec:further_analysis}. 

With the above assumption, we formalize the devised Retention Matrix $\mathbf{RM}^k$ for estimating the spectral norm of matrix $\mathbf{W}^k$ as: 
\begin{equation}
\begin{aligned}
     \mathbf{RM}^k &\triangleq \left[(\mathbf{A}^{k-1})^{\mathsf{T}} \mathbf{A}^{k}\right]^{\mathsf{T}} \left[(\mathbf{A}^{k-1})^{\mathsf{T}} \mathbf{A}^{k}\right]\\
     &=  (\mathbf{A}^{k-1})^{\mathsf{T}} (\mathbf{W}^k)^{\mathsf{T}} (\mathbf{A}^{k-1})
     (\mathbf{A}^{k-1})^{\mathsf{T}}  \mathbf{W}^k \mathbf{A}^{k-1}.\\
\end{aligned}
\label{eq:rm1}
\end{equation}

Incorporating independence assumption in Eq.~\ref{eq:9} (\textit{i.e.}, $(\mathbf{A}^{k-1})(\mathbf{A}^{k-1})=\mathbf{I}$)) with Eq.~\ref{eq:rm1}, we can transfer the $\mathbf{RM}^k$ as follows:
\begin{equation}
\label{eq:main}
     \mathbf{RM}^k = (\mathbf{A}^{k-1})^{\mathsf{T}} ({\mathbf{W}^k}^{\mathsf{T}}\mathbf{W}^k) \mathbf{A}^{k-1}.
\end{equation}
Based on Theorem 1 in supplemental material and Eq.~\ref{eq:main},  $\sigma_1(\mathbf{RM}^k) = \sigma_1({\mathbf{W}^k}^{\mathsf{T}}\mathbf{W}^k)$ where $\sigma_1(\cdot)$ is the function for computing the largest eigenvalue, \textit{i.e.}, Retention Matrix $\mathbf{RM}^k$ has the same largest eigenvalue with ${\mathbf{W}^k}^{\mathsf{T}}\mathbf{W}^k$. Thus, with the definition of spectral norm $\Vert \mathbf{W}^k\Vert_{SN} =  \sigma_1({\mathbf{W}^k}^{\mathsf{T}}\mathbf{W}^k)$, the spectral norm of the matrix $\mathbf{W}^k$ can be yielded through calculating the largest eigenvalue of $\mathbf{RM}^k$, \textit{i.e.}~$\sigma_1(\mathbf{RM}^k)$, which is solvable~\cite{shang2021lipschitz}. 

For networks with more complex layers, such as the residual block and block in MobileNet~\cite{he2016deep,howard2017mobilenets}, we can also design such a Retention Matrix to bypass the Lipschitz constant computation layer-wisely. By considering the block as an affine mapping from front to back activations, the proposed Retention Matrix can also be produced block-wisely, making our spectral norm calculation more efficient. Specifically, we define the Retention Matrix $\mathbf{RM}$ for the residual blocks as follows:
\begin{equation}
     \mathbf{RM}_m \triangleq \left[(\mathbf{A}^{f})^{\mathsf{T}} \mathbf{A}^{l}\right]^{\mathsf{T}} \left[(\mathbf{A}^{f})^{\mathsf{T}} \mathbf{A}^{l}\right],
\end{equation}
where $\mathbf{A}^{f}$ and $\mathbf{A}^{l}$ denote the front-layer activation maps and the back-layer activation maps of the residual block, respectively.
% \vspace*{-3mm}

\noindent\textbf{Calculation of Spectral Norms.} Here, to calculate the spectral norms of two matrices, an intuitive way is to use SVD to compute the spectral norm, which results in overloaded computation. Rather than SVD, we utilize \textbf{Power Iteration} method~\cite{golub2000eigenvalue,miyato2018spectral} to approximate the spectral norm of the targeted matrix with a small trade-off of accuracy. By Power Iteration Algorithm (see Supplemental Material), we can obtain the spectral norms of the binary and corresponding FP Retention Matrices, respectively (\textit{i.e.}~$\Vert \mathbf{RM}_F^k \Vert_{SN}$ and $\Vert \mathbf{RM}_B^k \Vert_{SN}$ for each $k\in \{1, \dots, L - 1\}$). And then, we can calculate the distance between these two spectral norms to construct the loss function.

\subsection{Binary Neural Network Optimization}
\label{sec:3.3}
\noindent\textbf{Optimization losses.} We define the Lipschitz continuity retention loss function $\mathcal{L}_{Lip}$ as
% \begin{equation}
%     \mathcal{L}_{Lip} = \sum_{k=1}^{L-1}(\frac{\Vert \mathbf{RM}_B^k \Vert_{SN} - \Vert \mathbf{RM}_F^k \Vert_{SN}}{\beta^{L-1-k}})^2,
%     \label{equ:15}
% \end{equation}
\begin{equation}
    \mathcal{L}_{Lip} = \sum_{k=1}^{L-1}\left[(\frac{\Vert \mathbf{RM}_B^k \Vert_{SN}}{\Vert \mathbf{RM}_F^k \Vert_{SN}} - 1){\beta^{k-L}}\right]^2,
    \label{equ:15}
\end{equation}
where $\beta$ is a coefficient greater than $1$. Hence, with $k$ increasing, the $\left[(\frac{\Vert \mathbf{RM}_B^k \Vert_{SN}}{\Vert \mathbf{RM}_F^k \Vert_{SN}} - 1){\beta^{k-L}}\right]^2$ increases. In this way, the spectral norm of latter layer can be more retained.

Combined with the cross entropy loss $\mathcal{L}_{CE}$, we propose a novel loss function for the overall optimization objective as
\begin{equation}
    \mathcal{L} = \frac{\lambda}{2}\cdot\mathcal{L}_{Lip} + \mathcal{L}_{CE},
    \label{equ:16}
\end{equation}
where $\lambda$ is used to control the degree of retaining the Lipschitz constant. We analyze the effect of the coefficient $\lambda$ in the supplementary material.
After we define the overall loss function, our method is finally formulated. 
The forward and backward propagation processes of \emph{LCR} are elaborated in Algorithm~\ref{alg:LCR}. 
% In practice, we follow the general BNN training methods proposed in IR-Net~\cite{qin2020forward} and ReActNet~\cite{liu2020reactnet} to train our binary models. 

\begin{algorithm}[!t]  
\caption{Forward and Backward Propagation of \emph{LCR}-BNN}  
\label{alg:LCR}
% \vspace*{-4mm}
% \begin{multicols}{2}
\begin{algorithmic}[1] 
	\Require A minibatch of data samples $(\mathbf{X,Y})$, current binary weight $\mathbf{W}_B^k$, latent full-precision weights $\mathbf{W}_F^k$, and learning rate $\eta$.
	\Ensure Update weights ${\mathbf{W}_F^k}^{\prime}$.
	\State \textbf{Forward Propagation}:
    \For{$k = 1$ to $L-1$}
\State Binarize latent weights: 
 $\mathbf{W}_B^k \xleftarrow{} \mathrm{sgn}(\mathbf{W}_F^k)$;
    	\State Perform binary operation with the activations of last layer: 
    	$\mathbf{A}_F^{k} \xleftarrow{} \mathbf{W}_B^k \cdot \mathbf{A}_B^{k-1}$;
    	\State Binarize activations: $\mathbf{A}_B^k \xleftarrow{} \text{sgn}(\mathbf{A}_F^k)$;
    	\State Produce the Retention Matrices $\mathbf{RM}_F^k$ and $\mathbf{RM}_B^k$ by Eq.~\ref{eq:main};	
    \EndFor
    \State Approximate the spectral norm of a series of $\mathbf{RM}$s by Algorithm~\ref{alg:pi} in the Supplemental Material, and obtain $\Vert \mathbf{RM}_F^k \Vert_{SN}$ and $\Vert \mathbf{RM}_B^k \Vert_{SN}$ for each $k \in \{1,\dots,L-1\}$;
    \State Compute the Lipschitz continuity retention loss $\mathcal{L}_{Lip}$ by Eq.~\ref{equ:15};
    \State Combine the cross entropy loss $\mathcal{L}_{CE}$ and the quantization error loss $\mathcal{L}_{QE}$ for the overall loss $\mathcal{L}$ by Eq.~\ref{equ:16};
	\State \textbf{Backward Propagation}:
    compute the gradient of the overall loss function, \textit{i.e.}~$\frac{\partial\mathcal{L}}{\partial \mathbf{W_B}}$, using the straight through estimator (STE)~\cite{bengio2013estimating} to tackle the sign function;
	\State \textbf{Parameter Update}:~update the full-precision weights: ${\mathbf{W}_F^k}^{\prime} \xleftarrow{} \mathbf{W}_F^k - \eta \frac{\partial\mathcal{L}}{\partial \mathbf{W}_B^k}$.
\end{algorithmic}  
\end{algorithm}

\noindent\textbf{Gradient Approximation.}
Several works~\cite{santurkar2018does,lin2019defensive,miyato2018spectral} investigate the robustness of neural networks by introducing the concept of Lipschitzness. In this section, we differentiate the loss function of our proposed method, and reveal the mechanism of how Lipschitzness effect the robustness of BNNs.

The derivative of the loss function $\mathcal{L}$ w.r.t $\mathbf{W}_B^k$ is:
\begin{equation}
\begin{aligned}
    &\frac{\partial\mathcal{L}}{\partial \mathbf{W}_B}
    = \frac{\partial (\mathcal{L}_{CE})}{\partial \mathbf{W}_B} + \frac{\partial (\mathcal{L}_{Lip})}{\partial \mathbf{W}_B^k}\\
    &\approx \mathbf{M} - \lambda\sum_{k=1}^{L-1}\beta^{k-L}(\frac{\Vert \mathbf{RM}_F^k \Vert_{SN}}{\Vert \mathbf{RM}_B^k \Vert_{SN}}) \mathbf{u}_1^k (\mathbf{v}_1^k)^{\mathsf{T}},\\
\end{aligned}
\label{eq:18}
\end{equation}
where $\mathbf{M} \triangleq \frac{\partial (\mathcal{L}_{CE})}{\partial \mathbf{W}_B}$, $ \mathbf{u}_1^k$ and $ \mathbf{v}_1^k$ are respectively the first left and right singular vectors of $\mathbf{W}_B^k$. In the content of SVD, $\mathbf{W}_B^k$ can be re-constructed by a series of singular vector, \textit{i.e.}
\begin{equation}
    \mathbf{W}_B^k = \sum_{j=1}^{d_k} \sigma_j(\mathbf{W}_B^k) \mathbf{u}_j^k \mathbf{v}_j^k,
    \label{eq:19}
\end{equation}
where $d_k$ is the rank of $ \mathbf{W}_B^k$, $\sigma_j(\mathbf{W}_B^k)$ is the $j$-th biggest singular value, $\mathbf{u}_j^k$ and $\mathbf{v}_j^k$ are left and singular vectors, respectively~\cite{shang2021lipschitz}. In Eq. \ref{eq:18}, the first term $\mathbf{M}$ is the same as the derivative of the loss function of general binarization method with reducing quantization error. As for the second term, based on Eq. \ref{eq:19}, it can be seen as the regularization term penalizing the general binarization loss with an adaptive regularization coefficient $\gamma \triangleq \lambda\beta^{k-L}(\frac{\Vert \mathbf{RM}_F^k \Vert_{SN}}{\Vert \mathbf{RM}_B^k \Vert_{SN}})$ (More detailed derivation can be found in the supplemental materials). 
Note that even we analyze the regularization property under the concept of SVD, we do not actually use SVD in our algorithm. And Eq.~\ref{eq:18} and~\ref{eq:19} only demonstrate that \emph{LCR} regularization is related to the biggest singular value and its corresponding singular vectors. The \emph{LCR} Algorithm~\ref{alg:LCR} only uses the Power Iteration (see Algorithm in the Supplemental Materials) within less iteration steps (5 in practice) to approximate the biggest singular value.

\noindent\textbf{Discussion on Retention Matrix}. Here, we would like to give a straight-forward explanation of why optimizing \emph{LCR} Loss in Eq.~\ref{equ:15} is equivalent to retaining Lipschitz continuity of BNN. 
Since the Lipschitz constant of a network $\Vert f\Vert_{Lip}$ can be upper-bounded by a set of spectral norms of weight matrices, \textit{i.e.} $\{\Vert \mathbf{W}_F^k \Vert_{SN}\}$ (see Eq.~\ref{eq:3}-\ref{eq:7}), we aim at retaining the spectral norms of binary weight matrices, instead of targeting on the network itself. And because Eq.~\ref{eq:9} to~\ref{eq:main} derive $\Vert \mathbf{RM}_F^k \Vert_{SN} = \Vert \mathbf{W}_F^k \Vert_{SN}$ and $\Vert \mathbf{RM}_B^k \Vert_{SN} = \Vert \mathbf{W}_B^k \Vert_{SN}$, we only need to calculate the spectral norm of our designed Retention Matrix $\Vert \mathbf{RM}_B^k \Vert_{SN}$. Finally, minimizing Eq.~\ref{equ:15} equals to enforcing $\Vert \mathbf{RM}_B^k \Vert_{SN} \longrightarrow \Vert \mathbf{RM}_F^k \Vert_{SN}$, which retains the spectral norm (Lipschitz continuity) of BNN. Therefore, the BNNs trained by our method have better performance, because the Lipschitz continuity is retained, which can smooth the BNNs.

\noindent \textbf{Differences with Spectral Normalization (SN) and Defensive Quantization (DQ).}
\label{sec:difference}
There are two major differences: (i) In contrast to SN and DQ directly calculating the spectral norm with weight matrix, our method compute the spectral norm of specifically designed Retention Matrix to approximate the targeted spectral norms by leveraging the activations in BNNs. In this way, we can approximate the targeted yet inaccessible Lipschitz constant of binary networks as shown in Fig.~\ref{fig:pipeline} (a), in which the weight matrix is extremely sparse. Particularly, instead of layer-wisely calculating the spectral norm of weight matrix proposed in SN, our method does \textit{not rely on weight matrix} since the calculation can be done using only the in/out activations (Eq.~\ref{eq:rm1}). (ii) To tackle the training architecture complexity, our designed Retention Matrix gives flexibility to regularize BNNs via utilizing Lipschitz constant in a module manner (\textit{e.g.}, residual blocks in ResNet~\cite{he2016deep}), instead of calculating the spectral norm and normalizing the weight matrix to 1 for each layer as shown in Fig.~\ref{fig:pipeline} (b). Benefit from  module-wise simplification, total computation cost of our method is much lower compared with SN and DQ.

\section{Experiments}
\label{sec:exp}
In this section, we conduct experiments on the image classification. Following popular setting in most studies\cite{qin2020forward,lin2020rotated}, we use the CIFAR-10~\cite{krizhevsky2012imagenet} and the ImageNet ILSVRC-2012~\cite{krizhevsky2012imagenet} to validate the effectiveness of our proposed binarization method. In addition to comparing our method with the state-of-the-art methods, we design a series of ablative studies to verify the effectiveness of our proposed regularization technique. All experiments are implemented using PyTorch \cite{paszke2019pytorch}. We use one NVIDIA GeForce 3090 GPU when training on the CIFAR-10 dataset, and four GPUs on the ImageNet dataset.

\noindent \textbf{Experimental Setup.} On CIFAR-10, the BNNs are trained for 400 epochs, batch size is 128 and initial learning rate is 0.1. We use SGD optimizer with the momentum of 0.9, and set weight decay is 1e-4. On ImageNet, the binary models are trained the for 120 epochs with a batch size of 256. We use cosine learning rate scheduler, and the learning rate is initially set to 0.1. All the training and testing settings follow the codebases of IR-Net~\cite{qin2020forward} and RBNN~\cite{lin2020rotated}.

\subsection{CIFAR}
CIFAR-10~\cite{krizhevsky2009learning} is  the  most  widely-used  image  classification dataset, which consists of 50K training images and 10K testing images of size 32×32 divided into 10 classes. For training, 10,000 training images are randomly sampled for validation and the rest images are for training. Data augmentation strategy includes random crop and random flipping as in~\cite{he2016deep} during training. For testing, we evaluate the single view of the original image for fair comparison.

\begin{table}[!t]
% \vspace{-0.4in}
\begin{minipage}{.49\textwidth}
    \centering
    % \captionsetup{font=small}
    \caption{Top-1 and Top-5 accuracy on ImageNet. ${\dagger}$ represents the architecture which varies from the standard ResNet architecture but in the same FLOPs level.}
    \scalebox{0.80}{
    \begin{tabular}{ccccc}\toprule
            \multirow{2}{*}{Topology} & \multirow{2}{*}{Method}  & BW & Top-1 & Top-5 \\ 
                                      &                          & (W/A)     & (\%)  & (\%) \\\midrule
                      & Baseline & 32/32           & 69.6       & 89.2       \\
                      & ABC-Net~\cite{lin2017towards} & 1/1             & 42.7       & 67.6       \\
                      & XNOR-Net~\cite{rastegari2016xnor}     & 1/1             & 51.2       & 73.2       \\
                      & BNN+~\cite{darabi2018bnn+}   & 1/1             & 53.0       & 72.6       \\
                      & DoReFa~\cite{zhou2016dorefa}  & 1/2             & 53.4       & -          \\
                      & BiReal~\cite{liu2020bi}  & 1/1            & 56.4       & 79.5       \\
                      & XNOR++~\cite{bulat2019xnor}   & 1/1             & 57.1       & 79.9       \\
                      & IR-Net~\cite{qin2020forward}   & 1/1             & 58.1       & 80.0       \\
                      & ProxyBNN~\cite{he2020proxybnn} & 1/1             & 58.7       & 81.2       \\
            \multirow{2}*{ResNet-18} & Ours     & 1/1             & \textbf{59.6} & \textbf{81.6} \\ \cline{2-5} 
                      & Baseline & 32/32           & 69.6       & 89.2       \\
                      & SQ-BWN~\cite{dong2017learning} & 1/32            & 58.4       & 81.6       \\
                      & BWN~\cite{rastegari2016xnor} & 1/32            & 60.8       & 83.0       \\
                      & HWGQ~\cite{li2017performance}  & 1/32            & 61.3       & 83.2       \\
                      & SQ-TWN~\cite{dong2017learning} & 2/32            & 63.8       & 85.7       \\
                      & BWHN~\cite{hu2018hashing}  & 1/32            & 64.3       & 85.9       \\
                      & IR-Net~\cite{qin2020forward}   & 1/32            & 66.5       & 85.9       \\
                      & Ours     & 1/32            & \textbf{66.9}  & \textbf{86.4}  \\ \midrule
                      & Baseline & 32/32           & 73.3       & 91.3       \\
                      & ABC-Net~\cite{lin2017towards}  & 1/1             & 52.4       & 76.5       \\
            ResNet-34 & Bi-Real ~\cite{liu2020bi} & 1/1             & 62.2       & 83.9       \\
                      & IR-Net~\cite{qin2020forward}   & 1/1             & 62.9       & 84.1       \\
                      & ProxyBNN~\cite{he2020proxybnn} & 1/1             & 62.7       & 84.5       \\
                      & Ours     & 1/1             & \textbf{63.5}  & \textbf{84.6} \\\midrule
            Variant   & ReActNet${\dagger}$~\cite{liu2020reactnet} & 1/1             & 69.4       & 85.5       \\
              ResNet   & Ours${\dagger}$     & 1/1             & \textbf{69.8}  & \textbf{85.7} \\\bottomrule
        \end{tabular}}
    \label{tabel:cifar}
\end{minipage}
%\hfill
\begin{minipage}{.49\textwidth}
    \centering
    % \captionsetup{font=small}
    \caption{Top-1 accuracy (\%) on CIFAR-10 (C-10) test set. The higher the better. W/A denotes the bit number of weights/activations. }
    \scalebox{0.9}{
    \begin{tabular}{p{1.5cm}ccc}
        \toprule
        \multirow{2}{*}{Topology}  & \multirow{2}{*}{Method}    & Bit-width       & Acc.         \\
                             &                & (W/A)           & (\%)         \\ \midrule
                             & Baseline       & 32/32           & 93.0          \\
        \multirow{2}*{ResNet-18}& RAD~\cite{ding2019regularizing}  & 1/1             & 90.5          \\
                             & IR-Net~\cite{qin2020forward}  & 1/1             & 91.5          \\
                             & Ours           & 1/1             & \textbf{91.8}       \\ \hline
                             & Baseline             & 32/32           & 91.7          \\
                             & DoReFa~\cite{zhou2016dorefa} & 1/1             & 79.3          \\
                             & DSQ~\cite{gong2019differentiable} & 1/1             & 84.1          \\
                             & IR-Net~\cite{qin2020forward}  & 1/1      & 85.5          \\
                             & IR-bireal~\cite{qin2020forward}  & 1/1             & 86.5          \\
                             & LNS~\cite{han2020training} & 1/1             & 85.7          \\
                             & SLB~\cite{yang2020searching} & 1/1             & 85.5          \\
                            & Ours           & 1/1             & \textbf{86.0} \\ 
        \multirow{2}*{ResNet-20}& Ours-bireal    & 1/1             & \textbf{87.2} \\ \cline{2-4}
                             & Baseline       & 32/32           & 91.7          \\
                             & DoReFa~\cite{zhou2016dorefa} & 1/32            & 90.0          \\
                             & DSQ~\cite{gong2019differentiable} & 1/32            & 90.1          \\
                             & IR-Net~\cite{qin2020forward}  & 1/32            & 90.2          \\
                             & LNS~\cite{han2020training} & 1/32            & 90.8          \\
        \multicolumn{1}{l}{} & SLB~\cite{yang2020searching} & 1/32            & 90.6          \\
                             & Ours           & 1/32            & \textbf{91.2} \\ \bottomrule
    \end{tabular}}
    \label{tabel:imagenet}
\end{minipage}
\vspace{-0.1in}
\end{table}

For ResNet-18, we compare with RAD~\cite{ding2019regularizing} and IR-Net~\cite{qin2020forward}. For ResNet-34, we compare with LNS~\cite{han2020training} and SLB~\cite{yang2020searching}, \textit{etc}. As the Table~\ref{tabel:cifar} presented, our method constantly outperforms other methods. \emph{LCR}-BNN achieves 0.3\%, 0.7\% and 0.6\% performance improvement over ResNet-18, ResNet-20 and ResNet-20 (without binarizing activations), respectively. In addition, our method also validate the effectiveness of bi-real structure~\cite{liu2020bi}. When turning on the bi-real module, IR-Net achieves 1.0\% accuracy improvements yet our method improves 1.2\%. 

\subsection{ImageNet}
ImageNet~\cite{deng2009imagenet} is a larger dataset with 1.2 million training images and 50k validation images divided into 1,000 classes. ImageNet has greater diversity, and its image size is 469×387 (average). The commonly used data augmentation strategy including random crop
and flipping in PyTorch examples ~\cite{paszke2019pytorch} is adopted for training. We report the single-crop evaluation result using 224×224 center crop from images. 

For ResNet-18, we compare our method with XNOR-Net~\cite{rastegari2016xnor}, ABC-Net~\cite{lin2017towards}, DoReFa~\cite{zhou2016dorefa}, BiReal~\cite{liu2020bi}, XNOR++~\cite{bulat2019xnor}, IR-Net~\cite{qin2020forward}, ProxyBNN~\cite{he2020proxybnn}. For ResNet-34, we compare our method with ABC-Net~\cite{lin2017towards}, BiReal~\cite{liu2020bi}, IR-Net~\cite{qin2020forward}, ProxyBNN~\cite{he2020proxybnn}. As demonstrated in Table~\ref{tabel:imagenet}, our proposed method also outperforms other methods in both top-1 and top-5 accuracy on the ImageNet. Particularly, \emph{LCR}-BNN achieves 0.9\% Top-1 accuracy improvement with ResNet-18 architecture, compared with STOA method ProxyBNN~\cite{he2020proxybnn}, as well as 0.6\% Top-1 accuracy improvement with ResNet-34 architecture, compared with state-of-the-art method ProxyBNN~\cite{qin2020forward}. Apart from those methods implemented on standard ResNet architectures, by adding our Lipschitz regularization module on ResNet-variant architecture, ReActNet~\cite{liu2020reactnet}, we also observe the accuracy improvement. Note that the training setting of adding our \emph{LCR} module on ReActNet is also different based on the codebase of ReActNet.

\subsection{Ablation Study} 
In this section, the ablation study is conducted on CIFAR-10 with ResNet-20 architecture and on ImageNet with ResNet-18. The results are presented in Table~\ref{tabel:ablation1}. By piling up our regularization term on IR-Net~\cite{qin2020forward} and ReActNet~\cite{liu2020reactnet}, our method achieves 1.2\% and 0.4\% improvement on ImageNet, respectively. Note that ReActNet is a strong baseline with a variant ResNet architecture.
We also study the effect of hyper-parameter $\lambda$ in loss function on CIFAR. As shown in Fig~\ref{tabel:ablation2}, we can observe that the performance improves with $\lambda$ increasing. Both experiments validate the effectiveness of our method. Apart from that, to investigate the regularization property of our method, we visualize several training and testing curves with various settings. Due to the space limitation, we put those demonstrations in the supplemental materials.

\begin{table}\footnotesize
\begin{minipage}[t]{.49\textwidth}
    \centering
    \caption{Effect of hyper-parameter $\lambda$ in loss function. Higher is better.}
        \scalebox{0.73}{
    \begin{tabular}{c|cccccc}
    \toprule
    \diagbox{Topology}{$\log_2\lambda$} & $\lambda=0$ & -1  & 0   & 1   & 2  & 3    \\ \specialrule{0.8pt}{0pt}{0pt}
    ResNet-18   & 85.9    & 86.2 & 87.9 & 90.1    & 91.2        &\textbf{91.8} \\
    ResNet-20   & 83.9    & 83.7 & 84.5 & 85.9    &\textbf{87.2}& 86.5  \\
    \bottomrule
    \end{tabular}}
    \label{tabel:ablation2}
    % \end{minipage}
    % \vfill
    % \begin{minipage}[t]{.49\textwidth}
    \vspace{0.1in}
    \centering
        \caption{Ablation Study of LCR-BNN.}
        \scalebox{0.83}{
        \begin{tabular}{clc}\toprule
        Dataset & Method         & Acc(\%) \\\hline
                                                  & Full Precision & 91.7    \\
                                                  & IR-Net~\cite{qin2020forward} (w/o BiReal)  & 85.5    \\
               CIFAR                              & IR-Net + LCR (w/o BiReal)  & 86.0    \\
                                                  & IR-Net~\cite{qin2020forward} (w/ BiReal)  & 86.5   \\
                                                  & IR-Net + LCR (w/o BiReal)   & 87.2     \\\midrule
                                                   & Full Precision & 69.6    \\
                                                  & IR-Net~\cite{qin2020forward} (w/o BiReal)  & 56.9    \\
         ImageNet                    & IR-Net + LCR (w/o BiReal)  &   58.4  \\
                                                  & IR-Net~\cite{qin2020forward} (w/ BiReal)  & 58.1   \\
                                                  & IR-Net + LCR    & 59.6    \\\cline{2-3}
                                                  & ReActNet & 69.4 \\
                                                  & ReActNet + LCR    & 69.8    \\
                                                  \bottomrule
        \end{tabular}}
        \label{tabel:ablation1}
    \end{minipage} 
    \hfill
    \begin{minipage}[t]{.49\textwidth}
        \centering
        \caption{FLOPS and BOPS for ResNet-18}
        \scalebox{0.86}{
        \begin{tabular}{ccc}\toprule
        Method         & BOPS                         & FLOPS                       \\\hline
        BNN~\cite{hubara2016binarized}  & $1.695\times{10}^9$          & $1.314\times{10}^8$         \\
        XNOR-Net~\cite{rastegari2016xnor}  & $1.695\times{10}^9$          & $1.333\times{10}^8$         \\
        ProxyBNN~\cite{he2020proxybnn} & $1.695\times{10}^9$          & $1.564\times{10}^8$         \\
        IR-Net~\cite{qin2020forward}  & $1.676\times{10}^9$          & $1.544\times{10}^8$         \\
        Ours           & $1.676\times{10}^9$          & $1.544\times{10}^8$         \\\hline
        Full Precision & $0$                          & $1.826\times{10}^9$         \\\bottomrule
        \end{tabular}}
        \label{tabel:cost}
    % \end{minipage}
    % \vfill
    % \begin{minipage}[t]{.19\textwidth}
    \vspace{0.25in}

\centering
        \captionof{table}{mCE on ImageNet-C. Lower is better.}
        \scalebox{0.9}{
        \begin{tabular}{cl}\toprule
        Method  & mCE (\%)\\\hline
        IR-Net~\cite{qin2020forward}  & 89.2 \\ 
        IR-Net + LCR (ours) & 84.9 $\downarrow$\\ \hline
        RBNN~\cite{lin2020rotated}  & 87.5 \\ 
        RBNN + LCR (ours) & 84.8 $\downarrow$\\ \hline
        ReActNet~\cite{liu2020reactnet}  & 87.0 \\ 
        IR-Net + LCR (ours) & 84.9 $\downarrow$\\ \toprule
        \end{tabular}}
        \label{table:imagenet-c}
    \end{minipage} 
% \vspace{-0.2in}    
\end{table}
\vspace{0.3in}

\subsection{Further Analysis}
\label{sec:further_analysis}
\noindent \textbf{Computational Cost Analysis.}
In Table~\ref{tabel:cost}, we separate the number of binary operations and floating point operations, including all types of operations such as skip structure, max pooling, \textit{etc}. It shows that our method leaves the number of BOPs and number of FLOPs constant in the model inference stage, even though our method is more computational expensive in the training stage. Thus, our Lipschitz regularization term does not undermine the main benefit of the network binarization, which is to speed up the inference of neural networks.

\begin{figure}[!t]
  \begin{center}
%   \vspace{-0.1in}
  \includegraphics[width=0.7\textwidth]{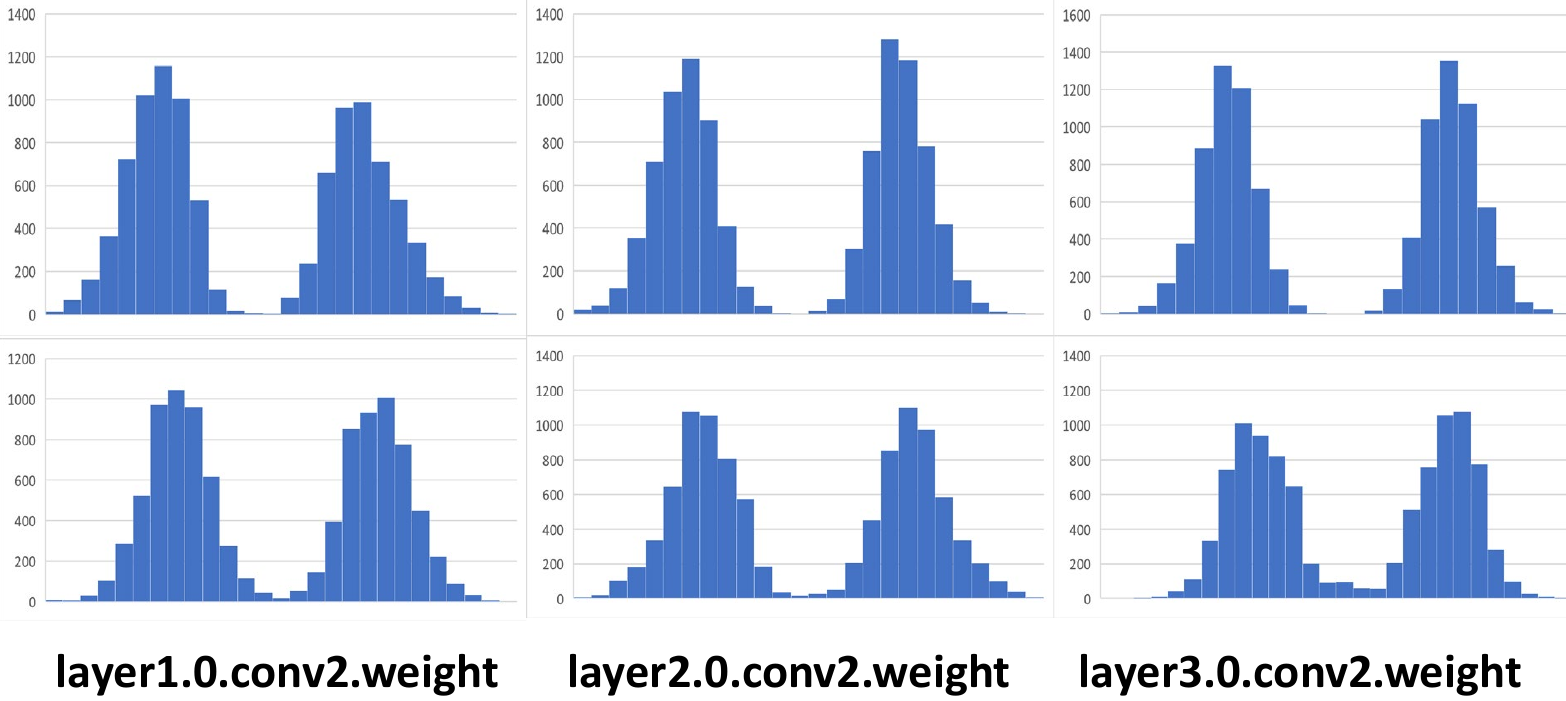}
  \end{center}
  \vspace{-0.2in}
  \caption{Histograms of weights (before binarization) of the IR-Net~\cite{qin2020forward} and \emph{LCR}-BNN with ResNet-18 architecture. The first row shows the results of the IR-Net, and the second row shows the results of ours. The BNN trained by our method has smoother weight distribution.}
\label{fig:dist}
\vspace{-0.2in}
\end{figure}

\noindent \textbf{Weight Distribution Visualization.} To validate the effectiveness of our proposed method from the perspective of weight distribution, we choose our \emph{LCR}-BNN and IR-Net to visualize the distribution of weights from different layers. For fair comparison, we randomly pick up 10,000 parameters in each layer to formulate the Figure~\ref{fig:dist}. Compared with IR-Net, the BNN trained by our method possesses smoother weight distribution, which correspondingly helps our method achieve 1.6\% accuracy improvement on ImageNet as listed in Table~\ref{tabel:imagenet}. More precisely, the standard deviation of the distribution of the IR-Net is 1.42, 28\% higher than ours 1.11, in the layer3.0.conv2 layer. 

\noindent \textbf{Robustness Study on ImageNet-C.} ImageNet-C~\cite{hendrycks2019benchmarking} becomes the standard dataset for investigation of model robustness, which consists of 19 different types of corruptions with five levels of severity from the noise, blur, weather and digital categories applied to the validation images of ImageNet (see Samples in Supplemental Materials). We consider all the 19 corruptions at the highest severity level (severity = 5) and report the mean top-1 accuracy. We use Mean Corruption Error (mCE) to measure the robustness of models on this dataset. We freeze the backbone for learning the representations of data \textit{w.r.t.} classification task, and only fine-tune the task-specific heads over the backbone (\textit{i.e.} linear protocol). The results in Table~\ref{table:imagenet-c} prove that add \emph{LCR} on the existing methods can improve the robustness of binary models.

\begin{figure}[!t]
  \begin{center}
    \includegraphics[width=0.85\textwidth]{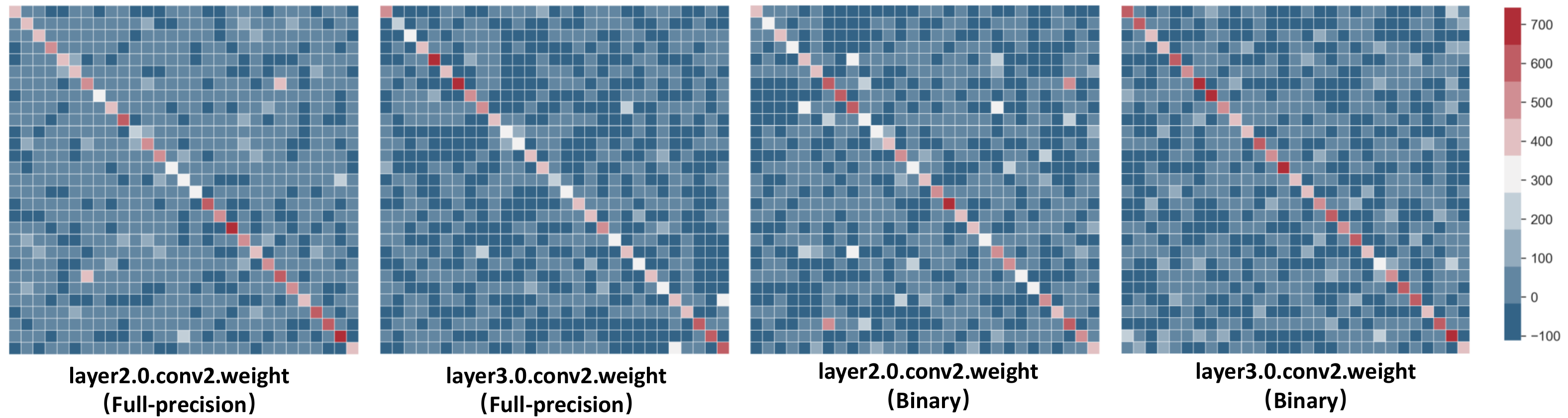}
  \end{center}
  \vspace{-0.2in}
  \caption{Correlation maps for reflecting independence assumption in Eq.~\ref{eq:9}.}
\label{fig:correlation}
\vspace{-0.2in}
\end{figure}

\noindent \textbf{Independence Assumption Reflection.} 
The assumption used in Eq.~\ref{eq:9} is the core of our method derivation, as it theoretically supports the approximation of the spectral norms of weight matrix with the designed retention matrix. Thus, we investigate this assumption by visualizing the correlation matrix of feature maps in the same batch.
Specifically, we visualise the correlation matrices of full-precision and binary activations, where red stands for two activations are similar and blue \textit{vice versa}. As shown in Fig~\ref{fig:correlation}, we can clearly observe that an activation is only correlated with itself, which largely testify this assumption. Besides, we also design another mechanism to use this assumption properly. We set a coefficient $\beta$ greater than $1$ to give more weight on latter layer's features such that they contribute more to $\mathcal{L}_{Lip}$ (Eq.~\ref{equ:15}). As in neural network, the feature maps of latter layers have stronger mutual linear independence~\cite{alain2016understanding}.

\section{Conclusion}
\label{sec:con}
In this paper, we introduce Lipschitz continuity to measure the robustness of BNN. Motivated by this, we propose \emph{LCR}-BNN to retain the Lipschitz constant serving as a regularization term to improve the robustness of binary models. Specifically, to bypass the NP-hard Lipschitz constant computation in BNN, we devise the Retention Matrices to approximate the Lipschitz constant, and then constrain the Lipschitz constants of those Retention Matrices. Experimental results demonstrate the efficacy of our method.

\noindent\textbf{Ethical Issues.} All datasets used in our paper are open-source datasets and do not contain any personally identifiable or sensitive personally identifiable information.
\noindent\textbf{Limitations.} Although our method achieve SoTA, adding it on existing method costs more time (around 20\% more) to train BNN, which is the obvious limitation of our method.

\noindent \textbf{Acknowledgements.} This research was partially supported by NSF CNS-1908658 (ZZ,YY), NeTS-2109982 (YY), Early Career Scheme of the Research Grants Council (RGC) of the Hong Kong SAR under grant No. 26202321 (DX), HKUST Startup Fund No. R9253 (DX) and the gift donation from Cisco (YY). This article solely reflects the opinions and conclusions of its authors and not the funding agents.

\clearpage
\bibliographystyle{splncs04}
\bibliography{egbib}

\clearpage
\setcounter{page}{1}
\section{Supplemental Material}
\label{sec:supp}

\subsection{Proofs.}

\noindent\textbf{Lemma 1.} If a function $f : \mathbb{R}^{n} \longmapsto \mathbb{R}^{m}$ is a locally Lipschitz continuous function, then $f$ is differentiable almost everywhere. Moreover, if $f$ is Lipschitz continuous, then
\begin{equation}
    \Vert f\Vert_{Lip} = \sup_{\mathbf{x}\in\mathbb{R}^{n}}\Vert  \nabla_{\mathbf{x}} f \Vert_2
\end{equation}
where $\Vert \cdot\Vert_2$ is the L2 matrix norm.

\noindent\textbf{Proof.} Based on Rademacher's theorem, for the functions restricted to some neighborhood around any point is Lipschitz, their Lipschitz constant can be calculated by their differential operator.

\noindent\textbf{Lemma 2.} 
Let $\mathbf{W} \in \mathbb{R}^{m \times n}, \mathbf{b} \in \mathbb{R}^{m}$ and $T(\mathbf{x}) = \mathbf{W}\mathbf{x} + \mathbf{b}$ be an linear function. Then for all $\mathbf{x} \in \mathbb{R}^{n}$, we have
\begin{equation}
    \nabla g(\mathbf{x}) = \mathbf{W}^{\mathsf{T}}\mathbf{W}\mathbf{x}
\end{equation}
where $g(\mathbf{x}) = \frac{1}{2}\Vert f(\mathbf{x}) - f(\mathbf{0})\Vert_2^2$.

\noindent\textbf{Proof.} By definition, $g(\mathbf{x}) = \frac{1}{2}\Vert f(\mathbf{x}) - f(\mathbf{0})\Vert_2^2 = \frac{1}{2}\Vert(\mathbf{W}\mathbf{x} + \mathbf{b}) - (\mathbf{W}\mathbf{0} + \mathbf{b})\Vert_2^2 = \frac{1}{2}\Vert \mathbf{W}\mathbf{x} \Vert_2^2$, and the derivative of this equation is the desired result.

\noindent\textbf{Theorem 1.} If a matrix $\mathbf{U}$ is an orthogonal matrix, such that $\mathbf{U}^\mathsf{T}\mathbf{U} = \mathbf{I}$, where $\mathbf{I}$ is a unit matrix, the largest eigenvalues of $\mathbf{U}^\mathsf{T} \mathbf{H} \mathbf{U}$ and $\mathbf{H}$ are equivalent:
\begin{equation}
    \sigma_1( \mathbf{U}^\mathsf{T} \mathbf{H} \mathbf{U}) = \sigma_1( \mathbf{H}),
\end{equation}
where the notation $\sigma_1(\cdot)$ indicates the largest eigenvalue of a matrix.

\noindent\textbf{Proof.} 
Because for $\mathbf{U}^{-1}$, we have 
\begin{equation}
    (\mathbf{U}^{-1})^\mathsf{T} (\mathbf{U}^\mathsf{T} \mathbf{H} \mathbf{U}) (\mathbf{U}^{-1}) = (\mathbf{U} \mathbf{U}^{-1})^\mathsf{T} \mathbf{H} (\mathbf{U} \mathbf{U}^{-1}) = \mathbf{H}.
\end{equation}
Thus matrix $(\mathbf{U}^\mathsf{T} \mathbf{H} \mathbf{U})$ and matrix $(\mathbf{H})$ are similar. The Theorem 1 can be proven by this matrix similarity.

\noindent\textbf{Exact Lipschitz constant computation is NP-Hard.} 
We take a 2-layer fully-connected neural network with ReLU activation function as an example to demonstrate that Lipschitz computation is not achievable in polynomial time. As we denoted in Method Section, this 2-layer fully-connected neural network can be represented as
\begin{equation}
     f(\mathbf{W}^1,\mathbf{W}^2;\mathbf{x})  = (\mathbf{W}^{2}\circ\sigma \circ\mathbf{W}^{1})(\mathbf{x}),
\end{equation}
where $\mathbf{W}^1\in\mathbb{R}^{d_0\times d_1}$ and $\mathbf{W}^2\in\mathbb{R}^{d_1\times d_2}$ are matrices of first and second layers of neural network, and $\sigma(x)=\max\{0, x\}$ is the ReLU activation function.

\noindent\textbf{Proof.} 
To prove that computing the exact Lipschitz constant of Networks is NP-hard, we only need to prove that deciding if the Lipschitz constant $\Vert f\Vert_{Lip} \leq L$ is NP-hard.

From a clearly NP-hard problem:
\begin{align}
     \label{eq:nphard}
     \max\min & \Sigma_i (\mathbf{h}_i^{\mathsf{T}} \mathbf{p})^2 = \mathbf{p}^{\mathsf{T}} \mathbf{H} \mathbf{p}\\
    & s.t.  \quad  \forall k, 0\leq p_k\leq1,
\end{align}
where matrix $\mathbf{H}=\Sigma_i \mathbf{h}_i \mathbf{h}_i^{\mathsf{T}}$ is positive semi-definite with full rank.
We denote matrices $W_1$ and $W_2$ as
\begin{equation}
    \mathbf{W}_1 = (\mathbf{h}_1, \mathbf{h}_2,\cdots,\mathbf{h}_{d_1}),
\end{equation}
\begin{equation}
    \mathbf{W}_2 = (\mathbf{1}_{d_1\times 1},\mathbf{0}_{d_1\times d_2 -1})^{\mathsf{T}},
\end{equation}
so that we have
\begin{equation}
    \mathbf{W}_2 \textnormal{diag}\left(\mathbf{p}\right) \mathbf{W}_1 =  \begin{bmatrix}
    \mathbf{h}_1^{\mathsf{T}} \mathbf{p} & 0 &  \dots & 0  \\
    \vdots & \vdots & \ddots &   \\
    \mathbf{h}_n^{\mathsf{T}} \mathbf{p}  & 0 &  &  0
    \end{bmatrix}^{\mathsf{T}}
\end{equation}

The spectral norm of this 1-rank matrix is $\Sigma_i (\mathbf{h}_i^{\mathsf{T}} \mathbf{p})^2$. We prove that Eq. \ref{eq:nphard} is equivalent to the following optimization problem
\begin{align}
\label{eq:nphard2}
     \max\min & \Vert \mathbf{W}_2 \textnormal{diag}\left(\mathbf{p}\right) \mathbf{W}_1 \Vert_2^2 \\
    & s.t. \quad   \mathbf{p} \in \left[0, 1\right]^n.
\end{align}
Because $H$ is full rank, $W_1$ is subjective and all $\mathbf{p}$ are admissible values for
$\nabla g(\mathbf{x})$ which is the equality case. Finally, ReLU activation units take their derivative within $\{0,1\}$ and Eq. \ref{eq:nphard2} is its relaxed optimization problem, that has the same optimum points. So that our desired problem is NP-hard.

\subsection{Power Iteration Algorithm} 
\begin{algorithm}[!h]  
	\caption{Compute Spectral Norm using Power Iteration}  
    % \vspace*{-5mm}
    \begin{multicols}{2}
	\begin{algorithmic}[1] 
		\Require Targeted matrix $\mathbf{RM}$ and stop condition $res_{stop}$.
		\Ensure The spectral norm of matrix $\mathbf{RM}$, \textit{i.e.},~$\Vert \mathbf{RM} \Vert_{SN}$.
		\State Initialize $\mathbf{v}_0 \in \mathbb{R}^m$ with a random vector.
	    \While{$res\geq res_{stop}$}
	    	\State $\mathbf{v}_{i+1} \gets \mathbf{RM}\mathbf{v}_{i} \bigl / \Vert \mathbf{RM}\mathbf{v}_{i}\Vert_2$
	    	\State $res = \Vert \mathbf{v}_{i+1} - \mathbf{v}_{i}\Vert_2$
		\EndWhile
		\State \Return{$\Vert \mathbf{RM} \Vert_{SN} = \mathbf{v}_{i+1}^{\mathsf{T}} \mathbf{RM} \mathbf{v}_{i} $}
	\end{algorithmic}
	\end{multicols}
	\label{alg:pi}
	\vspace*{-3mm}
\end{algorithm}

\subsection{Detailed derivation of the gradient.} 
The derivative of the loss function $\mathcal{L}$ w.r.t $\mathbf{W}_B^k$ is:
\begin{equation}
\begin{split}
    &\frac{\partial\mathcal{L}}{\partial \mathbf{W}_B}
    = \frac{\partial (\mathcal{L}_{CE})}{\partial \mathbf{W}_B} + \frac{\partial (\mathcal{L}_{Lip})}{\partial \mathbf{W}_B^k}\\
    &= \mathbf{M} - \lambda\sum_{k=1}^{L-1}\beta^{k-L}(\frac{\Vert \mathbf{RM}_F^k \Vert_{SN}}{\Vert \mathbf{RM}_B^k \Vert_{SN}})\frac{\partial \Vert \mathbf{RM}^k_B\Vert_{SN}}{\partial \mathbf{W}_B^k} \\
    &\approx \mathbf{M} - \lambda\sum_{k=1}^{L-1}\beta^{k-L}(\frac{\Vert \mathbf{RM}_F^k \Vert_{SN}}{\Vert \mathbf{RM}_B^k \Vert_{SN}})\frac{\partial \Vert \mathbf{W}^k_B\Vert_{SN}}{\partial \mathbf{W}_B^k}\\
    &\approx \mathbf{M} - \lambda\sum_{k=1}^{L-1}\beta^{k-L}(\frac{\Vert \mathbf{RM}_F^k \Vert_{SN}}{\Vert \mathbf{RM}_B^k \Vert_{SN}}) \mathbf{u}_1^k (\mathbf{v}_1^k)^{\mathsf{T}},\\
\end{split}
\label{eq:18}
\end{equation}
For the third equation:
\begin{equation}
        \mathbf{M} - \lambda\sum_{k=1}^{L-1}\beta^{k-L}(\frac{\Vert \mathbf{RM}_F^k \Vert_{SN}}{\Vert \mathbf{RM}_B^k \Vert_{SN}})\frac{\partial \Vert \mathbf{W}^k_B\Vert_{SN}}{\partial \mathbf{W}_B^k}
      \approx \mathbf{M} - \lambda\sum_{k=1}^{L-1}\beta^{k-L}(\frac{\Vert \mathbf{RM}_F^k \Vert_{SN}}{\Vert \mathbf{RM}_B^k \Vert_{SN}}) \mathbf{u}_1^k(\mathbf{v}_1^k)^{\mathsf{T}},
\end{equation}
we provide the core proof in here, \textit{i.e.} the first pair of left and right singular vectors of $\mathbf{W}_B$ can reconstruct $\frac{\partial \Vert \mathbf{W}_B\Vert_{SN}}{\partial \mathbf{W}_B} $ precisely. For $\mathbf{W}_B\in \mathbb{R}^{m\times n}$, the spectral norm $\Vert \mathbf{W}_B\Vert_{SN} = \sigma_1(\mathbf{W}_B)$ stands for its biggest singular value, $\mathbf{u}_1$ and $\mathbf{v}_1$ are correspondingly left and singular vectors. The SVD of $\mathbf{W}_B$ is $\mathbf{W}_B=\mathbf{U}\mathbf{\Sigma}\mathbf{V}^T$. Therefore $\Vert \mathbf{W}_B \Vert_{SN}=\mathbf{e}_1^T\mathbf{U}^T(\mathbf{U}\mathbf{\Sigma}\mathbf{V}^T)\mathbf{V}\mathbf{e}_1$, where $\mathbf{e}_1$ is the largest eigenvalue of matrix $\mathbf{W}_B^T\mathbf{W}_B$. Hence $\Vert \mathbf{W}_B \Vert_{SN} = \mathbf{u}_1^T\mathbf{W}_B\mathbf{v}_1$. Thus the derivative of spectral norm can be evaluated in the direction $\mathbf{H}$: $\frac{\partial \Vert W\Vert_{SN}}{\partial \mathbf{W}_B}(\mathbf{H}) = \mathbf{u}_1^T\mathbf{H}\mathbf{v}_1 = \mathrm{trace}(\mathbf{u}_1^T\mathbf{H}\mathbf{v}_1) = \mathrm{trace}(\mathbf{v}_1\mathbf{u}_1^T\mathbf{H})$. The gradient is $\frac{\partial \Vert \mathbf{W}_B\Vert_{SN}}{\partial \mathbf{W}_B}= \mathbf{v}_1\mathbf{u}_1^T$, which supports the Eq.13.

\subsection{ImageNet-C}
\begin{figure}[!h]
  \begin{center}
    \includegraphics[width=0.97\textwidth]{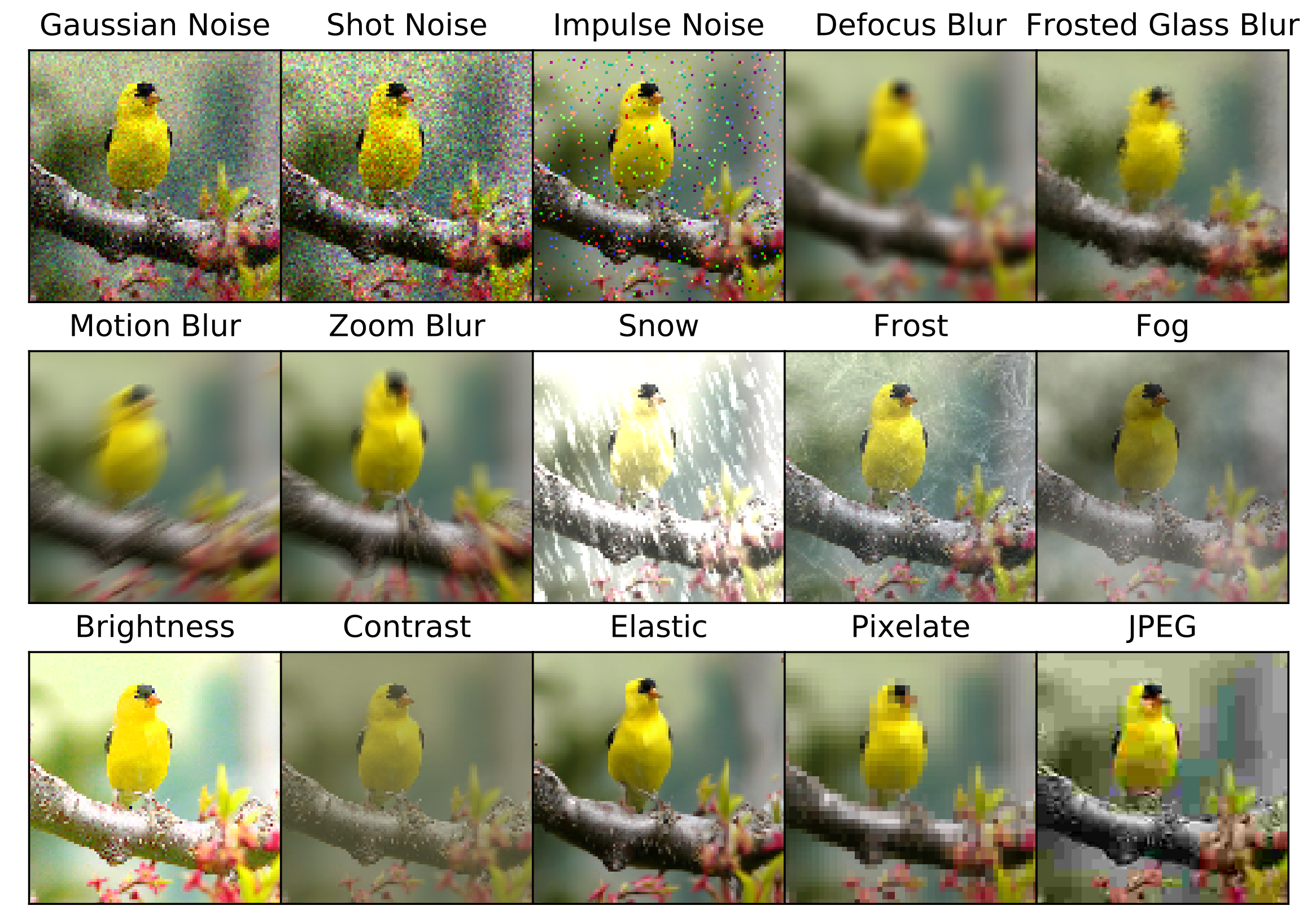}
  \end{center}
  \vspace{-0.3in}
  \caption{Examples of each corruption type in the image corruptions benchmark. While synthetic, this set of corruptions aims to represent natural factors of variation like noise, blur, weather, and digital imaging effects. This figure is reproduced from Hendrycks \& Dietterich (2019).}
\label{fig:imagenetc}
\vspace{-0.15in}
\end{figure}

\noindent\textbf{Sample Visualization of ImageNet-C.}
In Section 4.4 we evaluate methods on a common image corruptions benchmark (ImageNet-C) to demonstrate the effectiveness of \emph{LCR} from the perspective of model robustness. As illustrated in Section 4.4, ImageNet-C~\cite{hendrycks2019benchmarking} consists of 19 different types of corruptions with five levels of severity from the noise, blur, weather and digital categories applied to the validation images of ImageNet (see Fig.~\ref{fig:imagenetc}). As the figure presented, it is natural to introduce the ImageNet-C to measure the semantic robustness of models. Recently, ImageNet-C indeed has became the most widely acknowledged dataset for measuring the robustness of models.

\end{document}